\documentclass[runningheads]{llncs}
\usepackage[T1]{fontenc}
\usepackage{hyperref}
\usepackage{graphicx}
\usepackage{amsmath}
\usepackage{subcaption}
\usepackage{todonotes}
\usepackage{adjustbox}
\usepackage{booktabs}
\usepackage{enumitem}
\newcommand{\reduce}{\vspace{-0em}}
\newcommand{\reducesub}{\vspace{-0em}}
\newcommand{\reducebef}{\vspace{-0em}}

\newcommand{\compactsection}[1]{\section{#1}}
\newcommand{\compactsubsection}[1]{\subsection{#1}}

\begin{document}

\title{Assessing Trustworthiness of AI Training Dataset using Subjective Logic - A Use Case on Bias}
\author{K. Ismael Ouattara\inst{1,2}
\and
Ioannis Krontiris\inst{1}
\and
Theo Dimitrakos\inst{1}
\and
Frank Kargl \inst{2}
}


\institute{
Huawei Technologies, Munich, Germany \\
\email{\{koffi.ismael.ouattara, ioannis.krontiris, theo.dimitrakos\}@huawei.com}
\and
Ulm Universität, Ulm, Germany \\
\email{\{koffi.ouattara,frank.kargl\}@uni-ulm.de}
}

%
%

\maketitle
\vspace{-1em}
\begin{abstract}
As AI systems increasingly rely on training data, assessing dataset trustworthiness has become critical, particularly for properties like fairness or bias that emerge at the dataset level. Prior work has used Subjective Logic to assess trustworthiness of individual data, but not to evaluate trustworthiness properties that emerge only at the level of the dataset as a whole. This paper introduces the first formal framework for assessing the trustworthiness of AI training datasets, enabling uncertainty-aware evaluation of global properties such as bias. Built on Subjective Logic, our approach supports trust propositions and quantifies uncertainty in scenarios where evidence is incomplete, distributed, and/or conflicting. We instantiate this framework on the trustworthiness property of bias, and we experimentally evaluate it based on a traffic sign recognition dataset. The results demonstrate that our method captures class imbalance and remains interpretable and robust in both centralized and federated contexts.

\end{abstract}

\compactsection{Introduction}\label{sec:intro}
\reduce
Artificial Intelligence (AI) systems increasingly underpin high-stakes decisions in domains such as healthcare, finance, and autonomous mobility. While research has advanced the trustworthiness of AI models through fairness, explainability, and privacy-preserving techniques, less attention has been paid to the trustworthiness of the training datasets that underpin those models. Yet empirical studies show that dataset issues like sampling bias, label noise, and privacy vulnerabilities can undermine model fairness, robustness, and interpretability~\cite{schwabe2024metric}. In critical domains, even subtle flaws in data collection or curation can propagate into harmful downstream outcomes.

While data quality refers to intrinsic properties like accuracy, completeness, and consistency~\cite{ISO22989}, data trustworthiness is not a static attribute of data but a property that emerges from how data is produced, curated, and shared and how much confidence we can place in that process. A dataset may meet quality benchmarks yet remain untrustworthy, for example, if it originates from opaque or untrustworthy sources. For instance, in multi-agent systems such as connected vehicles, sensor data from different sources may individually appear high quality but differ in calibration, source reliability, or integrity~\cite{5GAA}. Similarly, in federated learning, local datasets may exhibit sampling bias or undetected distribution shifts, making global judgments difficult without modeling uncertainty in the evidence each node contributes. Trustworthiness thus builds upon data quality, adding an epistemic layer particularly relevant in decentralized or uncertain environments where evidence is fragmented or ambiguous.

There has been considerable progress so far in assessing and quantifying the trustworthiness of data based on actual evidence about various trust properties (e.g., integrity, authenticity, accuracy). Such frameworks, often grounded in models like Subjective Logic, enable reasoning under uncertainty, allowing systems to incorporate varying levels of belief and disbelief when forming trust evaluations. Rather than making assumptions, these systems continuously assess and re-evaluate trust using modular trust models and adaptable evaluation mechanisms.


However, these approaches largely focus on properties of individual data points, where evidence can often be tied to a specific source or transaction. What remains unsolved is how to assess dataset-level properties that only emerge when data is considered collectively. Properties like bias, fairness, or representativeness cannot be inferred from isolated records; they require a global view and sufficient coverage to be meaningfully evaluated. What kind of evidence do we need to evaluate such properties, and how do we quantify trust based on it, especially if such evidence is incomplete? Addressing this challenge calls for a new methodological foundation, one that enables reasoning under uncertainty, supports aggregation across evidence sources, and yields interpretable trust assessments for global dataset properties.

\emph{Contribution.}
This paper presents a novel methodology for assessing the trustworthiness of AI training datasets using Subjective Logic. Our approach systematically integrates uncertainty estimation and evidence collection into trust quantification. Our contributions are:
\vspace{-0.5em}
\begin{enumerate}
    \item A formal methodology for modeling dataset trustworthiness based on trust opinions, allowing properties such as fairness or bias to be expressed as trust propositions grounded in evidence;
    \item The introduction of two opinion quantification models under varying evidence conditions, ensuring the methodology remains robust and applicable across diverse dataset evaluation contexts;
    \item A concrete instantiation of the framework on the property of \textit{bias}, introducing two complementary trust opinion estimators, one based on class probability and one based on entropy, corresponding to both centralized and federated contexts;
    \item An empirical validation on a Traffic Sign Recognition dataset, demonstrating that the framework detects class imbalance, handles distributed evidence, and yields interpretable, uncertainty-aware trust assessments.
\end{enumerate}

The remainder of this paper is organized as follows. Section 2 reviews the foundations of Subjective Logic and key reasoning operators relevant to trust modeling. In Section 3, we discuss related work on dataset trustworthiness and bias quantification. Section 4 introduces our general framework for dataset or individual-level trustworthiness assessment, including trust proposition formulation and opinion quantification methods. Section 5 applies this framework to the property of bias, detailing two complementary evaluation methods. Section 6 presents experimental results using a traffic sign recognition dataset and analyzes the comparative behavior of both methods. Finally, Section 7 concludes the paper and outlines future research directions.
Additional diagrams, extended discussions, and implementation details are provided in the supplementary material, available at \url{https://github.com/Ouatt-Isma/-Trustworthiness-of-AI-Training-Dataset}.

\vspace{-1.5em}
\compactsection{Background on Subjective Logic}
\label{background}
Subjective Logic~(SL)~\cite{josangbook} is a probabilistic framework for reasoning under uncertainty, especially when information is incomplete or conflicting. It expresses a binomial opinion of an agent \(A\) on a binary variable \(X\) denoted by \(\omega_{X=x}^A~=~(b_x, d_x, u_x, a_x)\), where \(b_x\) is the belief that \(X = x\), \(d_x\) is the disbelief that \(X = x\) (i.e., belief in \(\bar{x}\)), \(u_x\) represents uncertainty (i.e., vacuity of evidence), and \(a_x\) is the base rate, a prior probability for \(X = x\) in the absence of evidence (typically \(0.5\)). These satisfy \(b_x + d_x + u_x = 1\).

SL enables the quantification of trust from positive \(r_x\) and negative \(s_x\) evidence using a prior weight \(W\) denoted by Baseline-Prior Quantification:
\begin{equation}
\label{eq:q1}
\begin{aligned}
    b_x &= \frac{r_x}{W + r_x + s_x}, \quad
    d_x = \frac{s_x}{W + r_x + s_x}, \quad
    u_x = \frac{W}{W + r_x + s_x}
\end{aligned}
\end{equation}
Note that in this quantification, uncertainty decreases
as the total number of positive and negative evidence increases.
In section~\ref{methodology:s}, we propose two other quantification models with different behavior.

SL also supports key reasoning operators such as Trust Discounting, Fusion, and logical operators~\cite{10706345,4622580}:

\begin{definition}[Logical Operators]
    \label{def:logicalops}
    Let \( A \) be an agent, and let \( X \) and \( Y \) be two random variables. Suppose that the analyst \( A \) holds an opinion \(\omega^A_{X = x}\) on the event \( X = x \) and \(\omega^A_{Y = y}\) on the event\( Y = y \). The opinion of analyst \( A \) on the compound statements \( (X = x) \wedge (Y = y) \) and \( (X = x) \vee (Y = y) \) can be computed using the following logical operators:
    \begin{align}
        \omega^A_{(X = x) \wedge (Y = y)} &= \omega^A_{X = x} \cdot \omega^A_{Y = y}, \notag  \\ 
        \omega^A_{(X = x) \vee (Y = y)} &= \omega^A_{X = x} \sqcup \omega^A_{Y = y}. \notag 
    \end{align}
    Moreover, for the negation operator, the opinion is simply inverted.
\end{definition}
    
\begin{definition}[Fusion]
    \label{def:fusion}
    Let \(A\) be an agent seeking to form an opinion about a variable \(X\). Assume that \(A\) has two independent sources or methods for deriving this opinion, denoted as \(P\) and \(Q\). The fused opinion, which combines the perspectives from \( P \) and \( Q \), is defined as:
    \begin{align}
        \omega^A_X = \omega^{P\& Q}_X = \omega^P_X \odot \omega^Q_X 
    \end{align}
    The choice of the fusion operator depends on the nature of the opinions being combined.
    \vspace{-1em}
    \paragraph{Example:} 
    Suppose we form an opinion based on images from two cameras and wish to fuse them. If the cameras share common angles, their opinions are correlated, suggesting that an averaging or weighted fusion method may be appropriate. Conversely, if the cameras capture entirely different angles, their opinions are independent, making cumulative fusion a more suitable approach.
\end{definition}
\vspace{-0.5em}
These mechanisms make Subjective Logic particularly suitable for evaluating dataset trust by integrating evidence with varying certainty and provenance.

\vspace{-1em}
\compactsection{Related Work}\label{rw}
\vspace{-1em}
Meyer et al.~\cite{meyer2023multiplicity} identify a key challenge in evaluating dataset trustworthiness: a single dataset can often support multiple valid interpretations of what it represents, due to ambiguity in how the dataset was constructed, e.g., unclear population coverage, subjective annotations, or undocumented filtering decisions. This insight emphasizes the epistemic uncertainty involved in dataset evaluation. Complementary efforts such as FRIES~\cite{rutinowski2024benchmarking} and METRIC~\cite{schwabe2024metric} attempt to systematize dataset evaluation through structured rubrics and expert-derived scores. However, these approaches presume complete and reliable metadata and offer no formal way to represent confidence or doubt in their assessments. Our work addresses this gap by formalizing trustworthiness as a probabilistic opinion over dataset properties using Subjective Logic, enabling assessments that reflect not only whether a dataset satisfies a given criterion, but also how uncertain that conclusion should be, based on the available evidence.

Focusing more specifically on dataset bias, traditional methods address class imbalance and demographic parity using statistical resampling or fairness metrics~\cite{10.1145/3600211.3604691,roy2022multi}. More recent work extends to multi-attribute fairness under imbalance~\cite{roy2022multi}, yet all these approaches assume that datasets are clean, fully labeled, and centralized. In federated and decentralized learning contexts, bias presents new challenges: sources such as client sampling heterogeneity, data partition skew, and limited metadata complicate fairness assessment~\cite{roy2024fairness,electronics13234664}. Kim et al.~\cite{electronics13234664} explore client-level bias mitigation (e.g., weighted aggregation, loss regularization) but do not provide methods to quantify trust in the bias assessment itself under uncertain or partial evidence. Our work fills this gap by offering a framework that quantifies trust in dataset bias (e.g., representativeness) even when data views are decentralized, incomplete, or noisy.

Subjective Logic (SL) has found diverse applications in AI systems for modeling uncertainty and trust. Prior work has applied SL to trust assessment in AI systems. DeepTrust~\cite{wang2020there} uses SL to evaluate neural network trustworthiness by incorporating algorithmic uncertainty, while recent efforts have extended SL to model evaluation in machine learning~\cite{10.1145/3605098.3635966,ouattara2025quantifying}, showcasing its utility in uncertainty-aware decision-making. More recently, Jake et al.~\cite{SLencodings} proposed Subjective Logic Encodings (SLEs), which model inter-annotator disagreement as probabilistic opinions to support data perspectivism in subjective labeling tasks. In general, these approaches focus on model-level trust or label-level uncertainty aggregation, without assessing dataset-level properties such as class imbalance or sampling bias. To our knowledge, this work is the first to apply SL to dataset trustworthiness assessment, capturing not only whether a dataset satisfies fairness-related criteria, but also the degree of confidence we can have given limited or noisy evidence. This shifts the use of SL from model-centric to data-centric trust reasoning.


\vspace{-0.5em}
\section{Proposed Framework for Dataset Trustworthiness}
\reduce
 \compactsubsection{Defining Trustworthiness of AI Training Dataset}\label{def}
\reducesub
We define the trustworthiness of a dataset as the degree to which a trustor (e.g., an AI developer or an automated system) can rely on the dataset to fulfill expectations relevant to a specific AI context. Formally, this trust is directional: an agent $A$ places trust in dataset $D$ to satisfy a trust property $P$. 

Trustworthiness can be assessed at two levels: individual entries and the dataset as a whole. At the entry level, key factors include reliability, source credibility, contextual relevance, and compliance with ethical and privacy standards. At the dataset level, trust involves fairness and representativeness (avoiding sampling and labeling biases), completeness and temporal relevance, privacy protections (e.g., differential privacy), resilience to adversarial or missing data, transparency through metadata, and provisions for human oversight.

In our framework, we express dataset trustworthiness as a trust opinion $\omega_D^A$ about whether a specific trust proposition related to $D$ is satisfied. These opinions are grounded in evidence, which may be partial, distributed, or referral-based. This definition supports modular trust assessment over dataset-level properties and enables uncertainty-aware reasoning, even in federated or noisy environments.

\reducebef
\subsection{Methodology for Dataset Trustworthiness Quantification }\label{methodology:s}
\reducesub


To assess dataset trustworthiness (at both the dataset and individual levels), we propose a high-level trust quantification methodology leveraging Subjective Logic, evidence-based theory, and trust modeling. This approach explicitly incorporates uncertainty, ensuring robust and context-aware trust assessments even in the presence of incomplete or conflicting evidence. Given its flexibility, many steps can be implemented in different ways. The process consists of:
\begin{enumerate}[nosep, leftmargin=*]
    \item Define the proposition to assess based on the trustworthiness property (e.g., \(B=\) the dataset is not biased).\label{methodology:s1}
    \item Express the proposition as a logical combination of atomic propositions that have quantifiable evidence. For example, if assessing trustworthiness of proposition \(B\), we can express it as: 
    \[ B = A \wedge R, \, \text{where} \wedge \, \text{is the logical AND operator} \] 
     where \(B\), \(A\), and \(R\) are respectively propositions "the dataset is not biased", "the dataset is accurate" and "the dataset is representative". If evidence suggests that \(A\) and \(R\) are not atomic, they can be further decomposed, or \(B\) can be decomposed in another way. 
     \label{methodology:s2}
    \item Identify trust sources for each atomic proposition~(as detailed in Section~\ref{sec:trustsources}) and collect evidence. A trust source refers to any entity, system, or mechanism that provides the basis for assessing trust. If evidence is not directly countable, an appropriate interpretation scheme must be devised to convert raw inputs into usable positive, negative, or uncertainty evidence. For instance, binary votes (e.g., approve/disapprove) provide directly countable evidence, while scalar ratings (e.g., on a 1–5 scale) require a mapping strategy to distinguish support, opposition, and uncertainty.\label{methodology:s3}
    \item Quantify the trust opinion for each trust source using the most appropriate quantification method. This choice depends on the characteristics of the available evidence (see discussion below). Then, combine them (this can be done using a fusion operator) and apply logical operators to evaluate the final opinion on the proposition. The choice of fusion operator depends on the dependencies between the trust sources, as explained in definition~\ref{def:fusion}.
    \label{methodology:s5}
\end{enumerate} 

In general, trust opinion quantification can be performed using different methods, each reflecting specific assumptions about uncertainty. In addition to the Baseline-Prior Quantification method presented in Section~\ref{background} (see Eq.~\ref{eq:q1}), we introduce two additional methods: Constant-Uncertainty Quantification, and Evidence-Weighted Quantification.
\begin{enumerate}
    
    \item Evidence-Weighted Quantification: introduces a weight \(w_x\) to account for uncertainty in addition to positive and negative evidence. It is particularly relevant when evidence itself is associated with an uncertainty score. The belief, disbelief, and uncertainty are computed as:
    \begin{equation}
    \label{eq:q3}
    \begin{aligned}
    \begin{cases}
        b_x &= \frac{r_x}{w_x + r_x + s_x}, \quad
        d_x = \frac{s_x}{w_x + r_x + s_x}, \\
        u_x &= \frac{w_x}{w_x + r_x + s_x},
        \end{cases}
    \end{aligned}
    \end{equation}

    \item Constant-Uncertainty Quantification: Unlike the two previous models, this method maintains a fixed uncertainty level \(U\), regardless of the number of evidence points. This is useful when uncertainty is predetermined by factors external to the number of positive and negative evidence. The belief and disbelief are scaled using a normalization factor (\(\gamma\)):
    \begin{align}
    \label{eq:q2}
    \begin{cases}
        u_x &= U,\quad
        \gamma = \frac{1-U}{r_x+s_x} \\
        b_x &=  \gamma\times r_x\quad
        d_x = \gamma\times s_x \\
        \end{cases}
    \end{align}
\end{enumerate}

Each of these methods provides a distinct approach to handling uncertainty, allowing flexibility depending on the available information and the desired level of confidence in the dataset trust assessment.

\compactsection{Application of the Methodology on Property of Bias}
In this section we apply the above general methodology on one specific trustworthiness property, namely dataset bias, and elaborate on the details of each of the steps. We chose bias because, unlike other properties such as data integrity or authenticity, it emerges only when the dataset is examined holistically, across distributions and demographic groupings.

\paragraph{Terminology.} We define bias as a systematic deviation in data representation that can lead to unfair, unreliable, or skewed model behavior, particularly by disadvantaging underrepresented groups~\cite{chouldechova2020snapshot}. Bias can stem from various factors, such as inadequate sampling, flawed data collection protocols, or subjective labeling. In our experiments, we operationalize bias through class imbalance, which is a measurable proxy defined by unequal label distributions, particularly relevant when richer demographic or contextual metadata is unavailable. 

To apply our methodology from Section~\ref{methodology:s}, we begin with Step~\ref{methodology:s1} by defining the trust proposition to be assessed: \( B = \text{"The dataset is not biased"} \). 

The following subsections walk through the remaining steps in this context. First, we examine how bias can be characterized and what types of evidence can support the trust proposition. We then detail how to identify trust sources, collect evidence, and quantify trust opinions. 

\vspace{-1em}
\compactsubsection{Bias Causes and Properties}

In our framework, the proposition “The dataset is not biased” is treated as a compound trust statement. We define this as the conjunction of two atomic propositions: (1) “The dataset is accurate” and (2) “The dataset is representative.” Indeed, bias in datasets can stem from two major dimensions: accuracy and representativeness. Accuracy bias arises when datasets misalign with true outcomes or labels, often due to label inconsistencies (label bias) or flawed data collection methods (measurement bias), both of which degrade model performance and reliability~\cite{gebru2018datasheets}. Representativeness bias, on the other hand, occurs when datasets fail to capture the diversity of the real world. This includes sampling bias, where certain groups are over- or underrepresented, as well as class imbalance, which impairs learning for minority groups. Temporal and geographic limitations further reduce generalizability, restricting the model’s effectiveness in broader contexts~\cite{barocas-hardt-narayanan}.



In conclusion, for Step~\ref{methodology:s2}, we can express: \[ B = A \wedge R, \, \wedge \, \text{is the logical AND operator} \] 
     where \(B\), \(A\), and \(R\) are, respectively, propositions "The dataset is not biased", "The dataset is accurate" and "The dataset is representative".


\subsection{Trust Sources for Bias}\label{sec:trustsources}

After expressing the proposition as a logical combination of
atomic propositions, we now identify the trust sources for each atomic proposition as described in Step~\ref{methodology:s3}. 
In this work, we propose the following trust sources based on the type of bias being quantified.
\paragraph{Measurement Bias (part of the accuracy property):}
Measurement bias occurs when the instruments or methods used for data collection systematically favor certain outcomes. Trust sources for evaluating measurement bias include anomaly detection in measurement tools, cross-validation using multiple devices, repeated measurements under similar conditions (self-consistency), and the precision level of the instruments as an indicator of accuracy.

\paragraph{Labeling Bias (part of the accuracy property):}
Labeling bias arises when labels in the dataset are applied inconsistently or are influenced by subjective or systemic factors. This bias can be quantified through inter-rater reliability, typically measured by metrics such as Cohen’s Kappa to assess annotator consistency.

\paragraph{Sampling Bias (part of the representativeness property):}
Sampling bias occurs when certain groups are disproportionately represented in the dataset. It can be evaluated by comparing the dataset’s sampling distribution to population-level data, analyzing over- or under-sampling patterns, and computing balance metrics such as entropy.

Now that we have listed the trust sources, still in Step~\ref{methodology:s3}, we have to collect evidence. In the following, we focus on quantifying the trust opinion for the balancing bias trust source only. In order to accommodate both centralized and collaborative dataset settings, and to illustrate the use of different trust quantification models, we introduce two methods. We will use these methods for collecting evidence (Step~\ref{methodology:s3}) and compute the opinion for this trust source only (step~\ref{methodology:s5}). 

\vspace{-1em}
\compactsubsection{Method 1: Class Probabilities-Based Quantification}
\vspace{-0.5em}
This method is an example of constant-uncertainty quantification because we argue that the uncertainty of the final opinion depends on the amount of data in the dataset. Here the uncertainty is fixed before processing the evidence. 

We introduce the concept of the \emph{distribution of class probabilities} to assess how well the dataset represents different classes. This method is particularly useful for understanding the uncertainty in class assignments within a dataset.
\vspace{-1em}
\subsubsection{Class Probabilities Distribution}
Given a classification task with \( K \) classes, we define the probability \( P(\text{Class}_k) \) that a random data point belongs to class \( k \) as:
\(
P(\text{Class}_k) = \frac{N_k}{N}
\)

where \( N_k \) is the number of data points belonging to class \( k \), and \( N \) is the total number of data points across all classes.

These probabilities form a \emph{probability distribution} over the classes, where:
\(
\sum_{k=1}^{K} P(\text{Class}_k) = 1
\)

This Class Probabilities Distribution concept is important because it helps us understand how data points are distributed across classes, allowing us to detect and quantify class imbalance.

To illustrate the impact of class distributions, we consider two dataset configurations:  
\begin{itemize}
    \item Imbalanced dataset (\( N = 5200 \)) with class counts \([100, 100, 1000, 1000, 1000, 2000]\), yielding probabilities \([0.019, 0.019, 0.192, 0.192, 0.192, 0.385]\).
    \item Balanced dataset (\( N = 5200 \)) with class counts \([857, 857, 867, 867, 876, 876]\), resulting in a more uniform probability distribution:\\ \([0.16481, 0.16481, 0.16673, 0.16673, 0.16846, 0.16846]\)
\end{itemize}
\begin{figure}[ht]
    \centering
    \captionsetup{justification=centering}
    \begin{subfigure}{0.4\textwidth}
        \centering
        \includegraphics[width=0.95\linewidth]{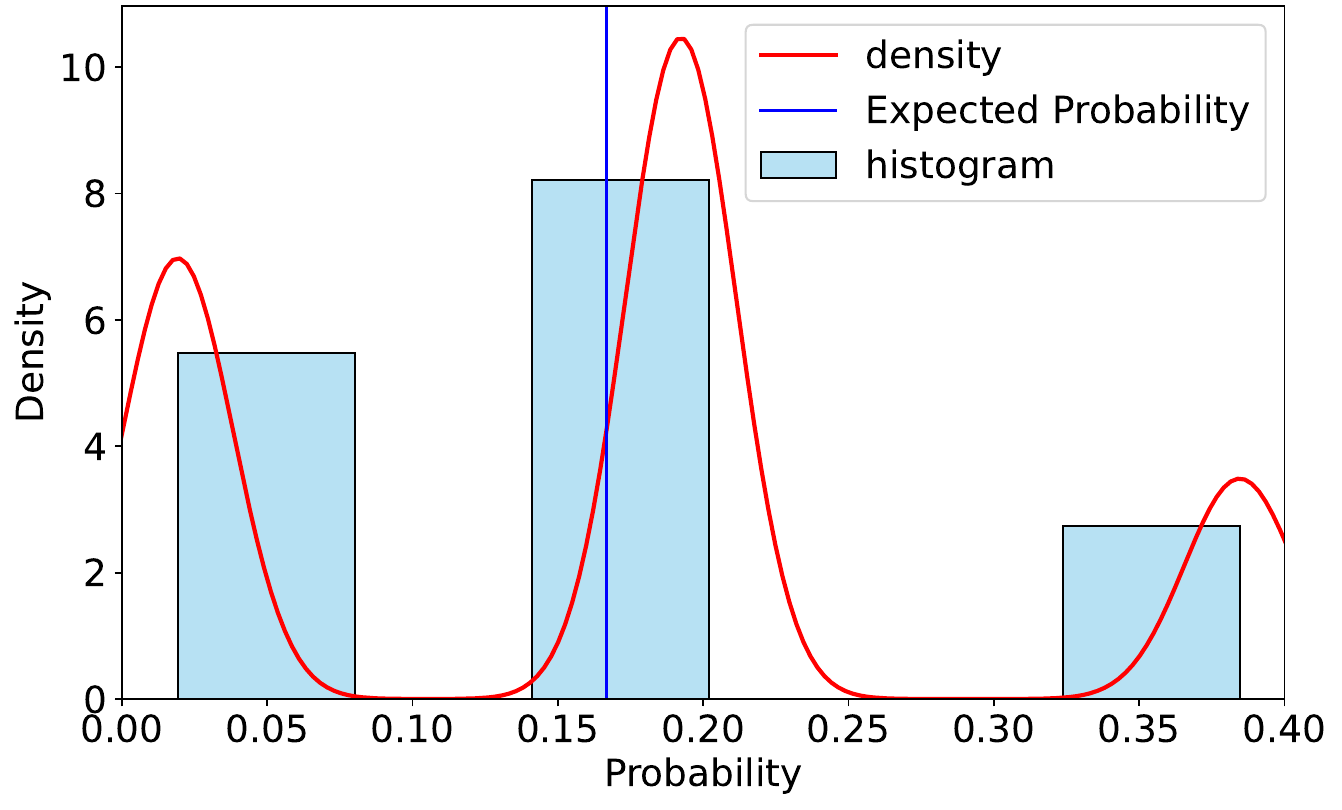}
        \caption{Example 1: imbalanced dataset}
    \end{subfigure}
    \begin{subfigure}{0.4\textwidth}
        
        \includegraphics[width=0.95\linewidth]{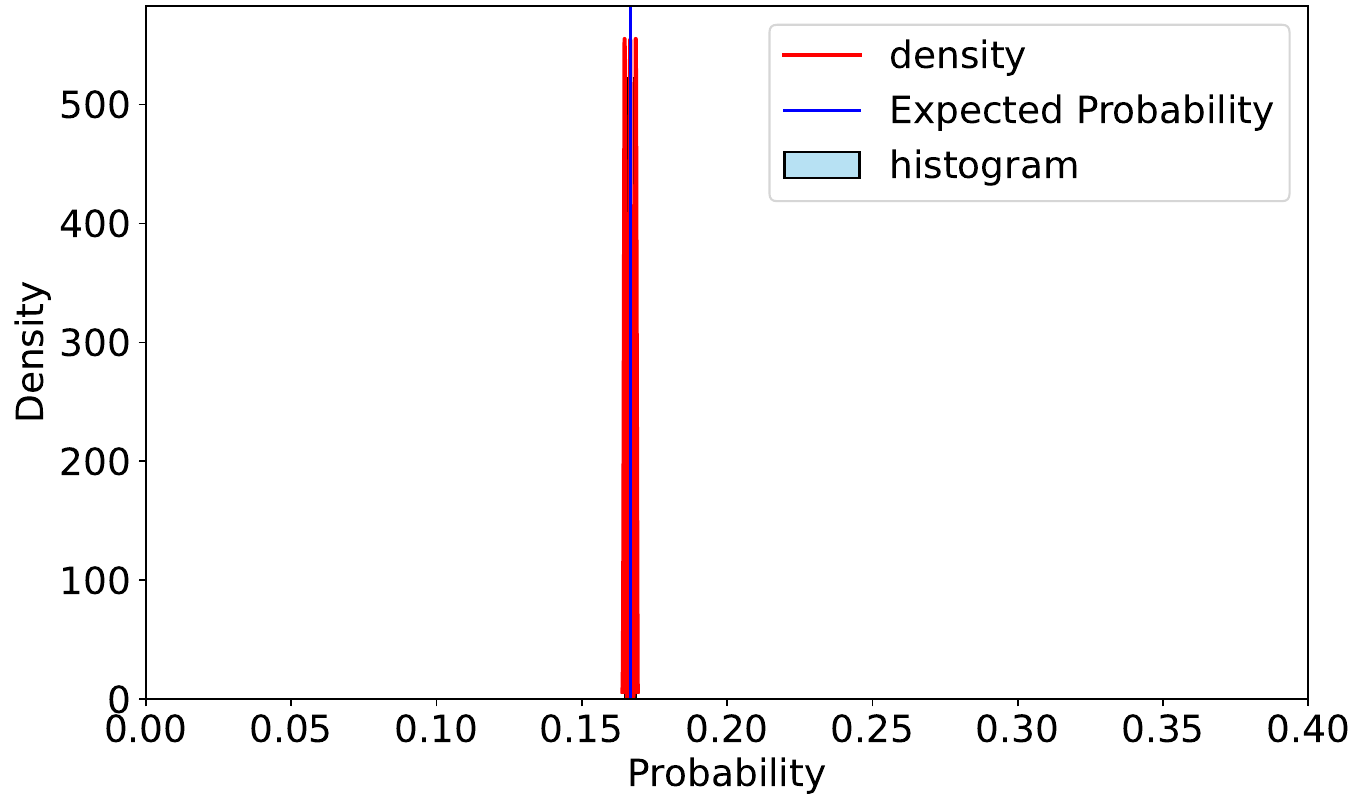}
        \centering
        \caption{Example 2: balanced dataset}
    \end{subfigure}
    \caption{Density plot of the class probabilities distribution.}
    \label{fig:probability-density}
\end{figure}
Figure \ref{fig:probability-density} visualizes the differences between these two cases. The imbalanced dataset exhibits a broader probability spread, while the balanced dataset’s distribution more closely resembles a Dirac delta function.
The imbalanced dataset shows significant class imbalance, potentially impacting
fairness and learning performance. In contrast, the more balanced dataset leads to a probability distribution closer to uniform,
improving class representation.
\vspace{-1em}
\subsubsection{Positive and Negative Evidence}
In a perfectly balanced dataset, all classes should have an equal probability of being selected. For \( n_c \) classes, the ideal probability for each class is \( \frac{1}{n_c} \). We compute the \emph{positive evidence} (\(r\)) by calculating the area under the density curve (of the Class Probabilities Distribution) within a tolerance zone, defined by \( [\frac{1}{n_c} - \eta_1, \frac{1}{n_c} + \eta_2] \). \(\eta_1, \eta_2\) are parameters used to increase or decrease the width of the tolerance zone.

The \emph{negative evidence} is simply \( s = 1 - r \) since the total area is 1 (as we have a probability density function). Regarding the uncertainty, we state that as the dataset grows, for a fixed Neural Network Model, the uncertainty decreases, and the trust opinion becomes more certain. We will see in the following paragraph how to calculate it. In summary:
\begin{align}
    r = \frac{n_t}{n_c} = \int_{\frac{1}{n_c} - \eta_1}^{\frac{1}{n_c} + \eta_2} f(x) \,dx , \; \text{and} \; s = 1 - r
\end{align}

where \(n_t\) is the number of classes within the tolerance zone and \(f\) is the density of the class probabilities distribution. For simplicity, one can assume a symmetric tolerance zone by setting \(\eta = \eta_1 = \eta_2\). Further details on the interpretation, selection, and impact of these parameters are provided in section~\ref{eval}.
\vspace{-1.5em}
\subsubsection{Uncertainty and Sample Complexity}
The uncertainty of the derived opinion can be quantified using the \emph{sample complexity} of the model and its \emph{VC dimension}. We use the result established by  Haussler et al.~\cite{HAUSSLER1994248}: 
\[\mathcal{M}(\epsilon, \delta) = \mathcal{O}\left( \frac{d}{\epsilon} \log\left(\frac{1}{\delta}\right) \right) \text{\cite{hanneke2016optimalsamplecomplexitypac}} \] Here, \(\epsilon\) represents the desired accuracy, and \(delta\) is the probability of failure. 

To quantify the uncertainty, we argue that when the dataset reaches a sufficient size, the uncertainty should be close or equal to zero. We use the following formula:

\begin{equation} \label{eq:uncertainty}
    U = \min \left( \max \left( \frac{\log(N_s) - \log(N)}{10}, m_1 \right), m_2 \right) 
\end{equation}

where \( N_s \) is the theoretical sample complexity set to \( \frac{d}{\epsilon} \log\left(\frac{1}{\delta}\right) \), \( N \) is the actual size of the dataset, \(m_1\) is the minimum desired uncertainty and \(m_2\) is the maximum desired uncertainty. For simplicity in the evaluation, we use \(m_1=0\) and \(m_2=1\).

This quantitative measure of the uncertainty is based on the dataset's size relative to the model's capacity. After getting the uncertainty~\eqref{eq:uncertainty}, one can compute the final opinion using Equation~\ref{eq:q2}.

\vspace{-1em}
\compactsubsection{Method 2: Entropy-Based Quantification}
\vspace{-0.5em}

\indent When dealing with multiple sub-datasets, a straightforward approach is to apply Method 1 and fuse all the resulting opinions. Additionally, we propose Method 2, an example of baseline-prior quantification~(see Equation~\ref{eq:q1}). 

In Method 2, we compute an entropy-based metric to assess the class balance of a sub-dataset. This value is then compared to a predefined threshold to determine whether to reinforce belief or disbelief in the dataset's trustworthiness. This method is suited for scenarios such as collaborative or federated learning, where data contributions come from multiple sources (e.g., OEMs).
\vspace{-1.5em}
\subsubsection{Entropy Computation}
Entropy is a measure of uncertainty in a probability distribution. Given a set of probabilities \( P = \{p_1, p_2, \dots, p_K\} \), the entropy \( \mathcal{H}(P) \) is defined as:
\(
\mathcal{H}(P) = -\sum_{k=1}^{K} p_k \log(p_k)
\)

In general, entropy helps quantify how "spread out" a probability distribution is. In our work, we use the entropy to understand how balanced the dataset is. Maximum entropy occurs when all classes are perfectly balanced. Conversely, minimum entropy occurs when all instances belong to a single class, indicating a completely imbalanced dataset. 

For our quantification, if the entropy exceeds a given threshold, it indicates \emph{positive evidence}; otherwise, it indicates \emph{negative evidence}. Next, we explain how to calculate this threshold.
\vspace{-1.5em}
\subsubsection{Threshold Calculation}
One way to determine a threshold is by computing the entropy for the \emph{edge-acceptable} case where probabilities are equally distributed within an acceptable tolerance zone, defined by:
\[
\text{ExpectedProb} \pm \eta
\]
Here, \( \text{ExpectedProb}=\frac{1}{n_c} \) and \(n_c\) is the number of classes.

This \emph{edge-acceptable} distribution is denoted as \(P_E\). The probabilities in this distribution are set to \(\text{ExpectedProb} + \eta\) or \(\text{ExpectedProb} - \eta\). If the number of classes is odd, we set one value to \(\text{ExpectedProb}\) to ensure that \(P_E\) sums to one.

The threshold \(T\) is then set to the entropy for this distribution:
\(
T = \mathcal{H}(P_E)
\).

Once the evidence is classified as positive or negative, we obtain a count of positive and negative evidence. Then we can derive the opinion using Equation~\ref{eq:q1}.

Note that Method 2 can be extended into an Evidence-Weighted Quantification where the uncertainty weight increases based on how far the computed entropy is from the threshold. If the evidence deviates significantly from the threshold, the associated uncertainty will be small.
\reducebef
\compactsection{Experimental Evaluation and Discussion}\label{eval}
\reduce
In this section, we experimentally evaluate our framework by applying it to the German Traffic Signs Dataset~\cite{GTSRB} using a Convolutional Neural Network (CNN). The dataset contains over 50,000 images across 43 traffic sign classes, with each image standardized to a resolution of \(32\times32\) pixels. This provides a realistic benchmark for evaluating computer vision classification systems under class imbalance scenarios.

For our experiment we define two scenarios. In Scenario 1, we consider a centralized dataset and evaluate two cases:
\begin{enumerate}
    \item \textbf{Original Dataset}: Using the original dataset without any modifications.
    \item \textbf{Augmented Dataset}: Applying data augmentation techniques to the original dataset to increase balance across classes.
\end{enumerate}
In Scenario 2, instead of relying on a single dataset, we assume that multiple Original Equipment Manufacturers (OEMs) contribute to building the dataset. This collaborative approach introduces new challenges, such as imbalanced datasets across contributors. We investigate two cases:

\begin{enumerate}
    \item \textbf{Original Dataset}: The original dataset is split into sub-datasets representing a dataset for each OEM.
    \item \textbf{Imbalanced Collaborative Dataset}: Some of the sub-datasets are intentionally imbalanced, simulating scenarios where certain data sources are biased. To create an imbalanced sub-dataset, we removed all warning signs from it.
\end{enumerate}

For both scenarios, we have \(\text{ExpectedProb}=\frac{1}{43}=0.02326\), and we set \(\eta = \eta_1=\eta_2=0.02\). For the experiment, we used a Convolutional Neural Network (CNN) trained on the German TSR Benchmark~\cite{GTSRB}. The model architecture includes three convolutional layers with 64, 128, and 256 filters (kernel size $3 \times 3$), followed by batch normalization, $2 \times 2$ max-pooling, and dropout (rate 0.25) after each block. Two fully connected layers with 256 and 128 units (ReLU activation) are followed by a final dropout (rate 0.5) and a softmax output layer with 43 units. The training was performed using the Adam optimizer, sparse categorical crossentropy loss, and accuracy as the evaluation metric. The model was trained for 5 epochs with a batch size of 36 using the Keras framework.

We evaluated the dataset’s trustworthiness under two different scenarios: a centralized dataset and a collaborative dataset contributed by multiple sources. In the following paragraphs, we discuss the obtained results. 

\begin{figure}[ht]
    \centering
    \captionsetup{justification=centering}
    \begin{subfigure}{0.35\textwidth}
 
        \centering
        \includegraphics[width=0.95\linewidth]{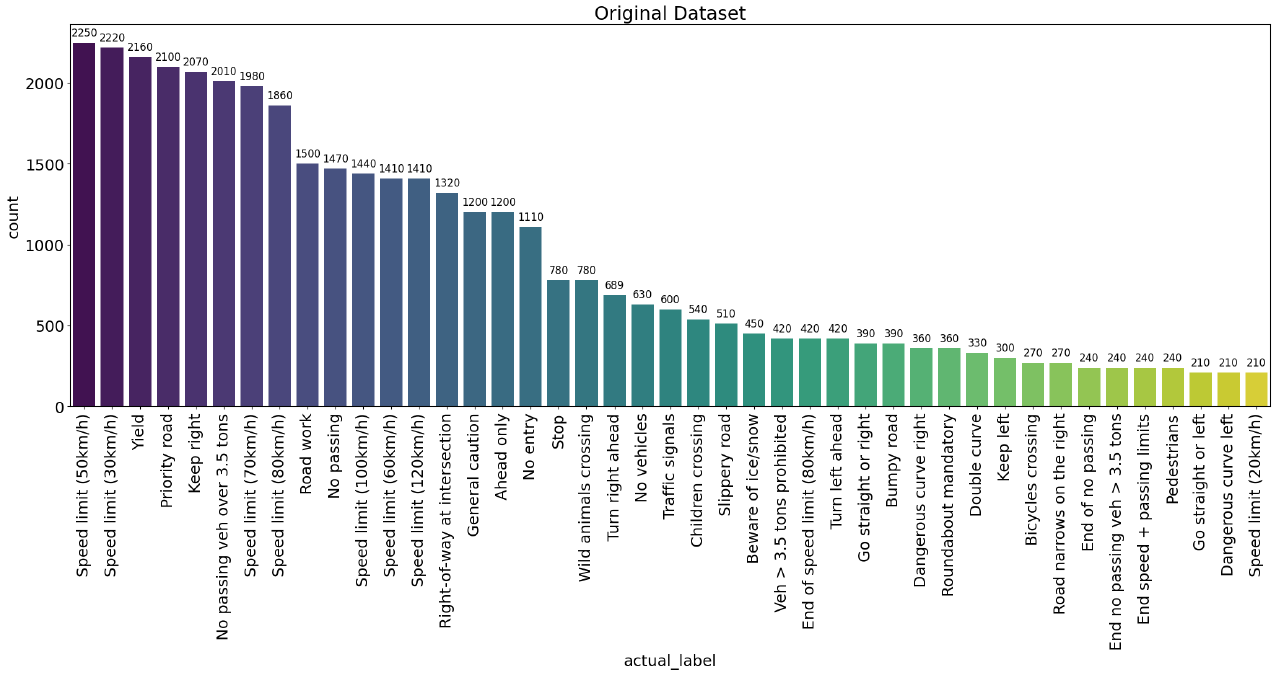}
        \caption{Distribution for original dataset}
    \end{subfigure}
    \begin{subfigure}{0.35\textwidth}
        
        \includegraphics[width=0.95\linewidth]{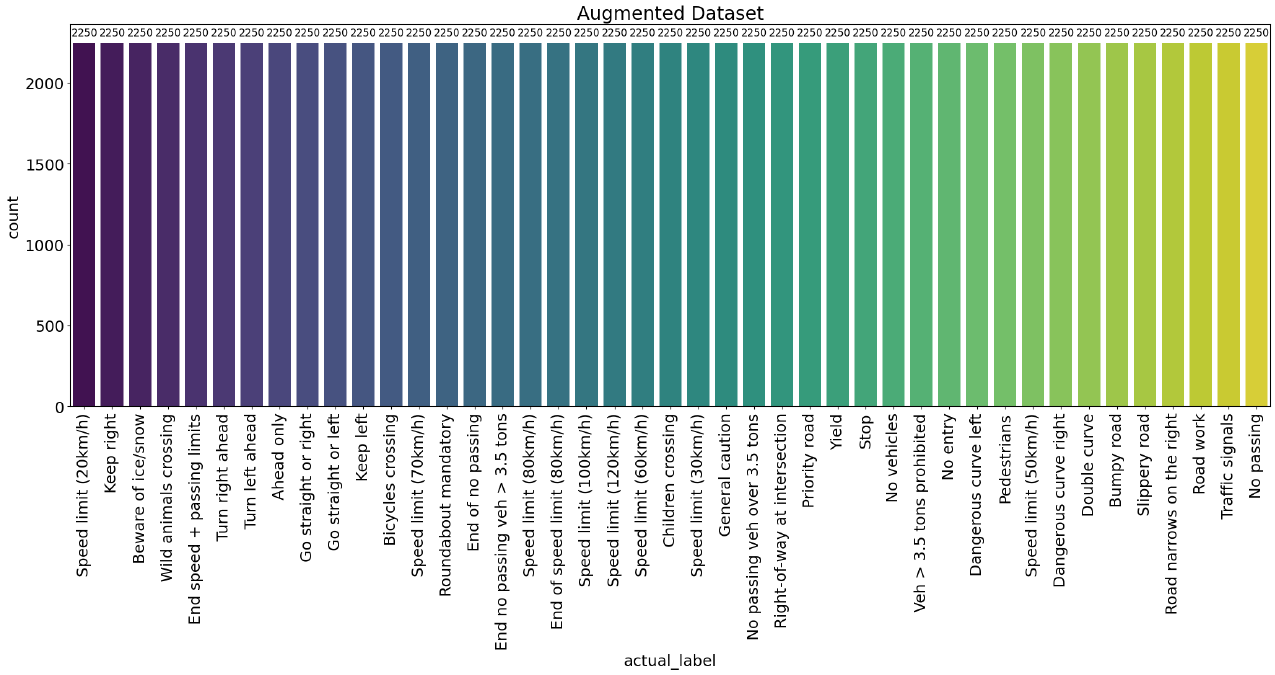}
        \centering
        \caption{Distribution for augmented dataset}
    \end{subfigure}
    \begin{subfigure}{0.35\textwidth}
        \centering
        \includegraphics[width=0.95\linewidth]{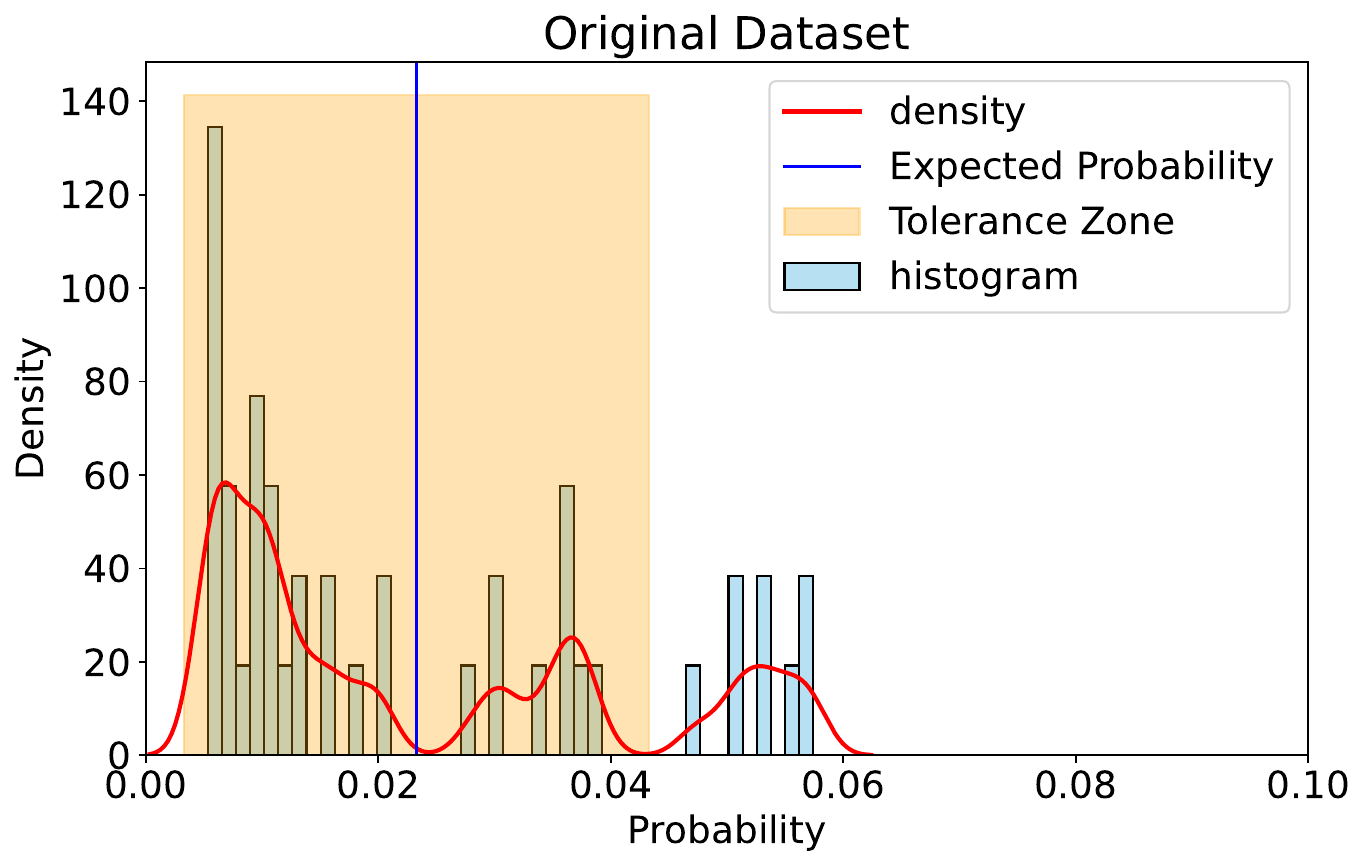}
        \caption{Probability Distribution for original dataset}
    \end{subfigure}
    \begin{subfigure}{0.35\textwidth}
    \includegraphics[width=0.95\linewidth]{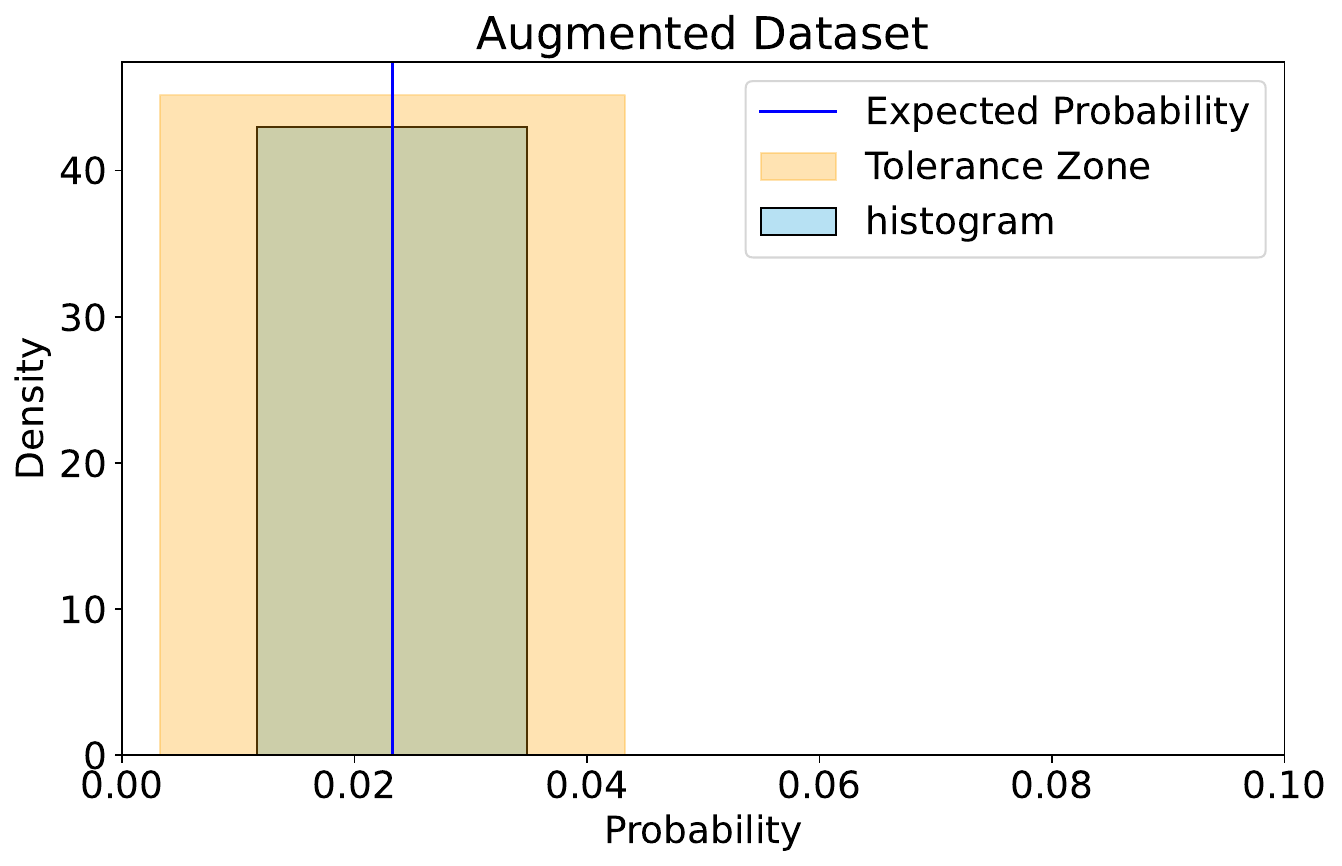}
        \centering
        \caption{Probability Distribution for augmented dataset}
    \end{subfigure}
    \caption{Class and probability distribution for scenario 1}
    \label{fig:distscenario1}
\end{figure}

In Scenario 1, we compared the original dataset with an augmented version\footnote{We applied SMOTE~\cite{DBLP:journals/corr/abs-1106-1813} designed to improve class balance to synthetically oversample minority classes by interpolating in feature space, then reshaped the output back into image form for training.}. The original dataset exhibited class imbalance, while augmentation significantly improved the uniformity of label distribution, as shown in Figure~\ref{fig:distscenario1}. This refinement was also reflected in the class probability density plots in Figure~\ref{fig:distscenario1} indicates a better alignment with the expected probability distribution (a narrower spread). Table~\ref{tab:acc1}  summarizes the performance results, showing that while overall accuracy remained nearly identical (95.17\% for the original dataset and 95.10\% for the augmented one), class-specific improvements were evident. Accuracy on warning signs\footnote{We report accuracy separately for warning signs because they were intentionally imbalanced in scenario 2 of our experiments. Other signs are grouped to serve as a reference for comparison.} increased from 95.19\% to 95.40\%, and for other signs, it improved from 89.95\% to 91.68\%, thereby reducing bias. 
The computed trust opinions also support this improvement, with belief increasing from 0.50 to 0.64 and disbelief dropping from 0.11 to 0.

\begin{figure}[ht]
    \centering
    \begin{subfigure}{0.35\textwidth}
        \includegraphics[width=0.95\linewidth]{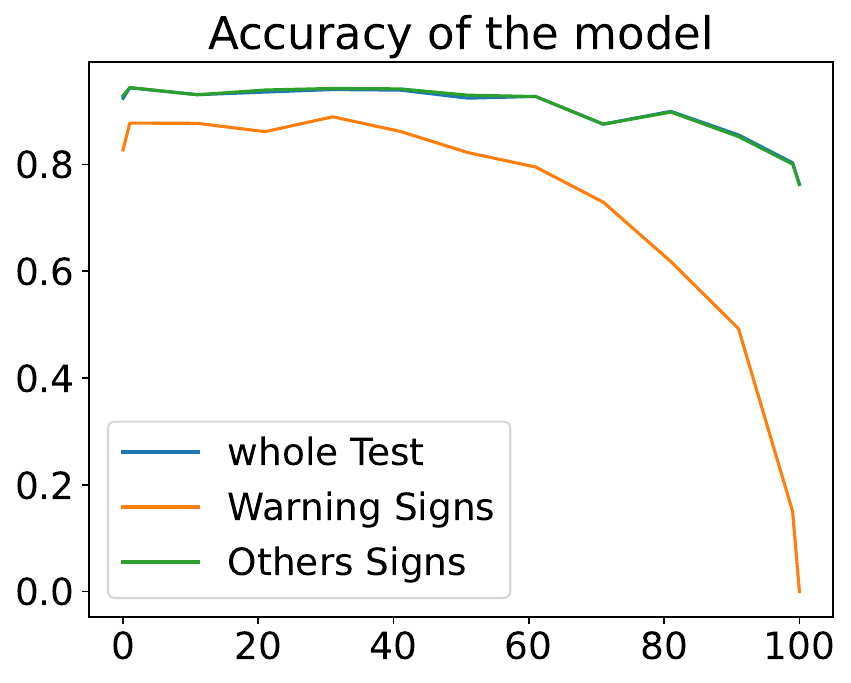}
        \caption{Accuracy}
    \end{subfigure}
    \begin{subfigure}{0.35\textwidth}
        \includegraphics[width=0.95\linewidth]{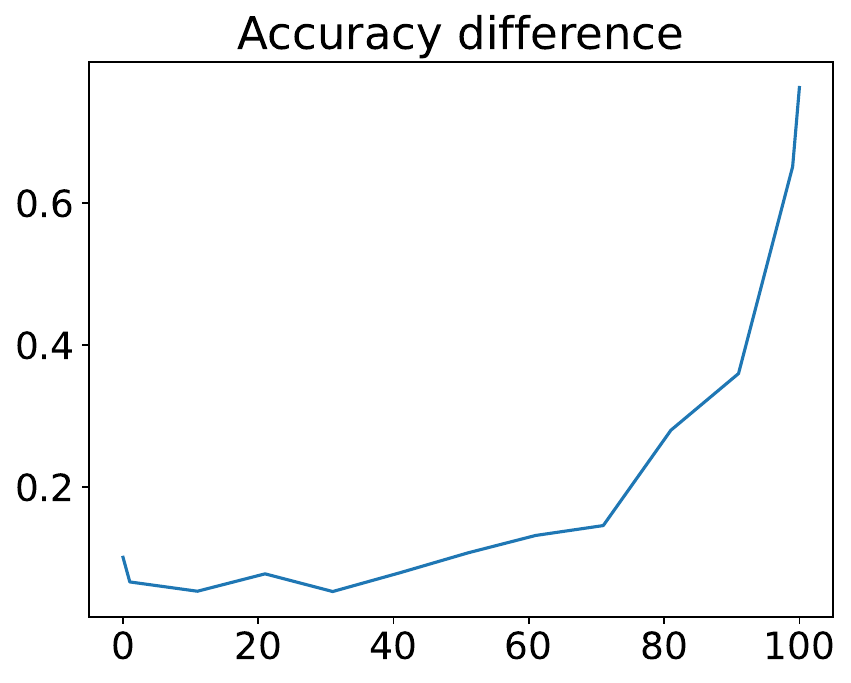}
        \caption{Accuracy difference}
    \end{subfigure}
    \caption{Plot of accuracy on warning signs and others vs number of imbalanced sub datasets (for 100 OEMs)}
    \label{fig:accplot}
\end{figure}

\begin{figure}[ht]
    \centering
    \begin{subfigure}{0.35\textwidth}        \includegraphics[width=0.95\linewidth]{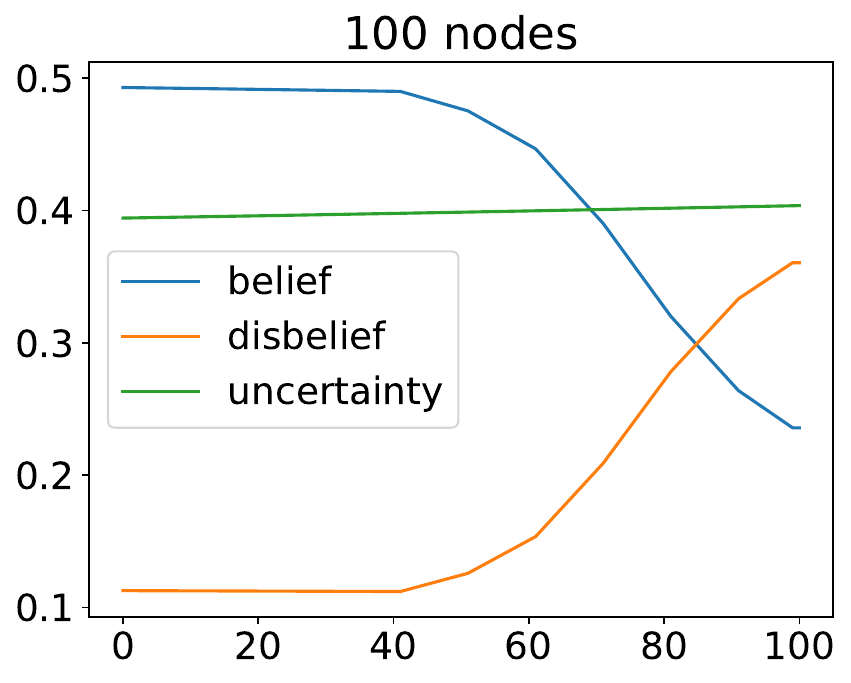}
        \caption{Method 1 when 100 OEMs}
    \end{subfigure}
    \begin{subfigure}{0.35\textwidth}
        \includegraphics[width=0.95\linewidth]{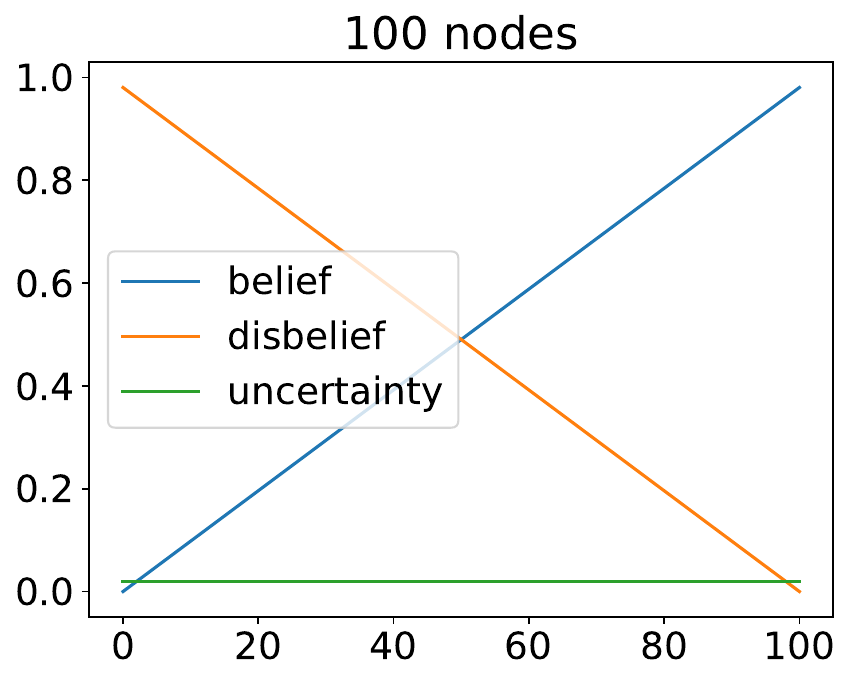}
        \caption{Method 2 when 100 OEMs}
    \end{subfigure}
    \begin{subfigure}{0.35\textwidth}
        \includegraphics[width=0.95\linewidth]{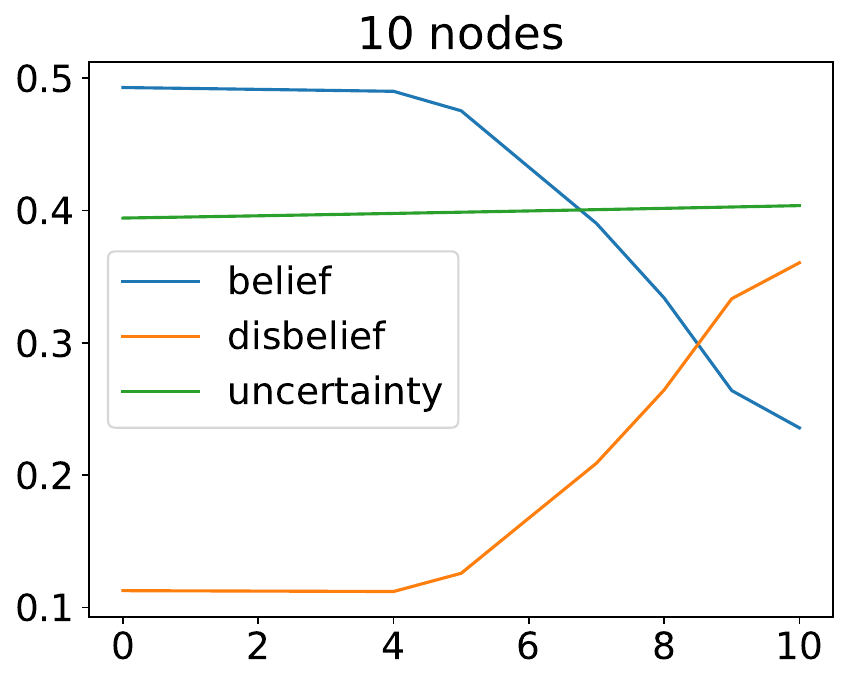}
        \caption{Method 1 when 10 OEMs}
    \end{subfigure}
    \begin{subfigure}{0.35\textwidth}
        \includegraphics[width=0.95\linewidth]{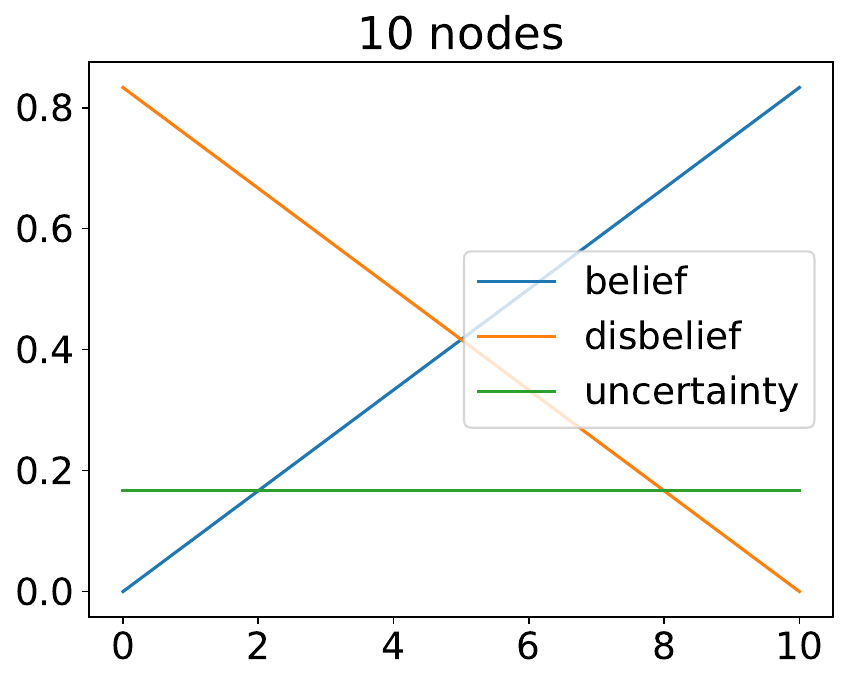}
        \caption{Method 2 when 10 OEMs}
    \end{subfigure}
    
    \caption{Method 1 and method 2 for 10 and 100 OEMs}
    \label{fig:op}
\end{figure}

In Scenario 2, we assessed the impact of imbalanced sub-datasets contributed by 100 OEMs. As shown in Figure~\ref{fig:accplot}, the model’s accuracy on warning signs dropped drastically, from 82\% to 0\%, while the performance on other signs decreased from 92\% to 76\%. This widening accuracy gap underscores the negative impact of imbalanced sub-datasets, which became more pronounced as their number increased. The trust opinion analysis, presented in Figure~\ref{fig:op}, highlights distinct behaviors of the two evaluation methods. For Method 1~\footnote{Method 1 applied on the obtained dataset after concatenation of all sub datasets}, belief initially decreased gradually but dropped sharply after approximately 60 imbalanced sub-datasets, while uncertainty remained stable at around 0.4, demonstrating its reliance on the total dataset size rather than its distribution across sources. Method 2 exhibited a linear decrease in belief and a symmetrical increase in disbelief, indicating a more deterministic confidence shift as dataset imbalance grew. When applied to datasets with only 10 OEMs instead of 100, Method 1 yielded similar belief and disbelief trends, since the total dataset remains the dominant factor. However, for Method 2, the uncertainty increased to approximately 0.2, reflecting the reduced number of evidence sources in a smaller dataset.

\begin{table}[]
\centering
\begin{adjustbox}{width=0.5\textwidth}
    \begin{tabular}{|c|c|c|}
    \hline
        Dataset &  Original & Augmented \\
        \hline
        \hline
         Accuracy & 95.17\% & 95.10\% \\
         Accuracy on Warning & 95.19\% & 95.4\% \\
         Accuracy on Other Signs & 89.95\% & 91.68\% \\
         Accuracy difference & 5.24\% & 3.88\% \\
        Trust Opinion & \( (0.5, 0.11, 0.39) \) & \((0.64, 0, 0.36)\) \\
       \hline
    \end{tabular}
\end{adjustbox}
\caption{Evaluation of both trained models}
\label{tab:acc1}
\end{table}
The results demonstrate fundamental differences in how Methods 1 and 2 quantify uncertainty. Method 1 maintains a fixed uncertainty level, making it well-suited for centralized settings where dataset size remains constant. Its behavior suggests that the method is relatively unaffected by the distribution of data across nodes, which is useful when assessing datasets that are static and fully known. However, this also limits its applicability in dynamic, distributed environments where data contributions vary.

Method 2, on the other hand, gradually reduces uncertainty as more sources contribute. This characteristic makes it more suitable for federated learning and other decentralized applications, where data arrives from multiple sources with varying degrees of completeness and bias. Unlike Method 1, Method 2 does not require access to the full label distribution, relying instead on entropy computations, which enhances its privacy-preserving properties.

A critical parameter in Method 1 is the \textit{tolerance} $\eta$, which defines the acceptable range around the expected class probability $\frac{1}{n_c}$ used to compute positive evidence. A class probability falling within the interval
\(
\left[\frac{1}{n_c} - \eta,\ \frac{1}{n_c} + \eta\right]
\)
is considered balanced. The choice of $\eta$ significantly affects the resulting belief, disbelief, and uncertainty. If $\eta$ is too small, even minor class imbalances are penalized, leading to high disbelief. If $\eta$ is too large, meaningful imbalance may go undetected.


\begin{figure}[htbp]
  \centering
  \begin{minipage}{0.65\textwidth}
    \includegraphics[width=\linewidth]{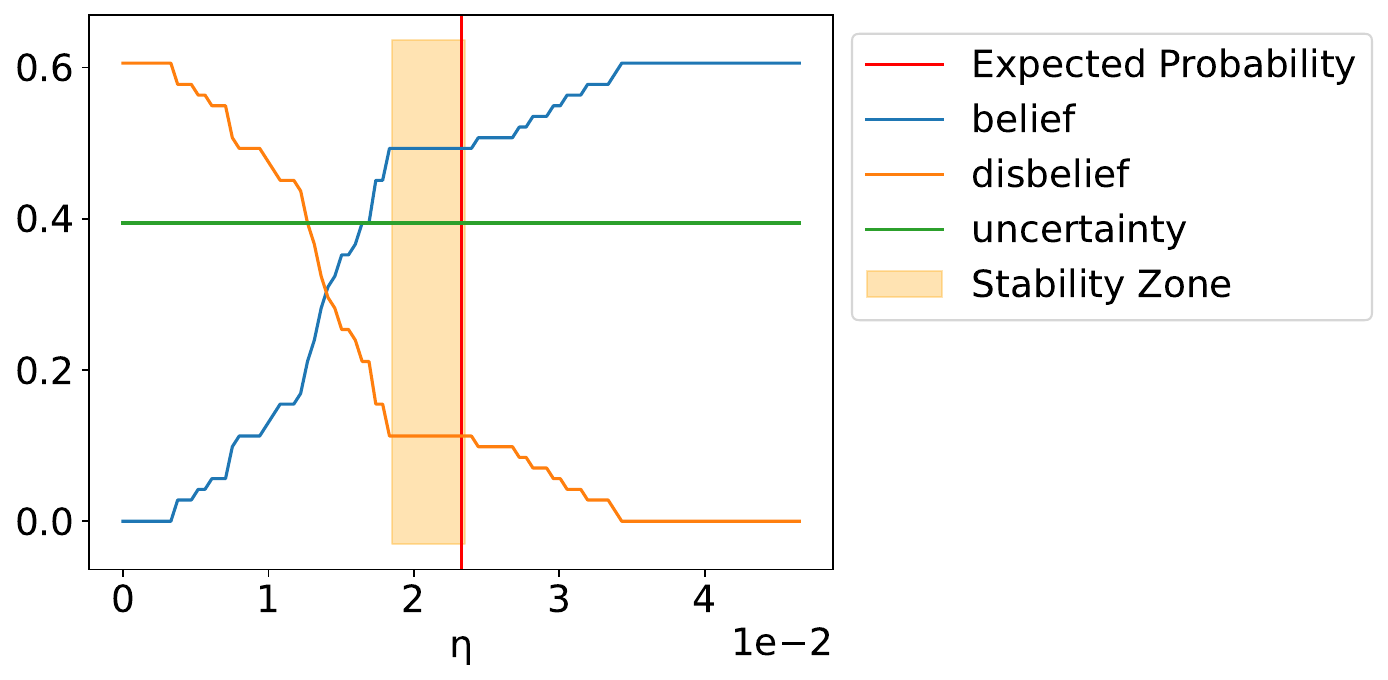}
  \end{minipage}%
  \hfill
  \begin{minipage}{0.3\textwidth}
    \vspace{0.2em}
    \caption{Effect of \(\eta\) on belief, disbelief, and uncertainty. The red vertical line represents the expected probability, located at \(\eta = 0.02326\).}
    \label{fig:eta_effect}
  \end{minipage}
  
  \vspace{0.5em} 
  The red vertical line represents the expected probability, located at $\eta = 0.02326$. The blue and orange curves depict the evolution of belief and disbelief, respectively, while the green curve represents uncertainty. The shaded region ($0.0185 \leq \eta \leq 0.0235$) indicates a \textbf{stability zone}, where belief and disbelief exhibit minimal fluctuation, suggesting a region of robust trust quantification. Beyond this range, belief stabilizes at a high value while disbelief approaches zero, reducing sensitivity to minor class imbalance. Uncertainty remains constant across $\eta$, reflecting a controlled trust assessment.
\end{figure}

Figure~\ref{fig:eta_effect} illustrates the behavior of belief, disbelief, and uncertainty as a function of $\eta$. Notably, belief increases and stabilizes as $\eta$ grows, while disbelief decreases. The uncertainty remains constant. Based on this behavior, we define a \textit{stability zone} for trust quantification: $0.0185 \leq \eta \leq 0.0235$. Within this range, trust scores show minimal fluctuation. We select $\eta = 0.02$ as it lies in the middle of this region and provides stable results.

These observations emphasize the importance of parameter selection when applying Method 1. Although the method is structurally robust, sensitivity to $\eta$ must be considered during evaluation.

\vspace{-1em}
\paragraph{Privacy Considerations.} Both methods support privacy-preserving evaluation. Method 1 relies on aggregated label distributions, avoiding access to individual data points. Method 2 goes further, using entropy-based metrics that eliminate the need for raw label access. These characteristics make both approaches suitable for trust assessment in privacy-sensitive contexts.

\vspace{-1em}
\paragraph{Imbalance as Contextual Property.}
In some applications, class imbalance reflects real-world distributions, and altering it could reduce model fidelity. In such cases, imbalance is not necessarily a flaw but a contextual feature.  Furthermore, in resource-constrained environments, replicating or augmenting minority classes may not be feasible. Interpreting trust scores should account for this low belief due to imbalance might be acceptable if the dataset aligns with expected deployment scenarios. A future extension could integrate a metric comparing dataset distribution to real-world class frequencies.

\vspace{-1em}
\compactsection{Conclusion} \label{conclusion}
\vspace{-1em}

This paper presents a structured methodology for quantifying AI dataset trustworthiness using Subjective Logic. Applied to a TSR use case, the approach identified class imbalances via probability and entropy-based methods. Results suggest Method 1 suits centralized datasets, while Method 2 is better adapted to decentralized privacy-sensitive settings. The framework enables quantifiable trust assessment, contributing to more transparent and fair AI systems.

Focusing on class imbalance as a measurable bias, this preliminary work primarily demonstrates the methodology. Future work will address other bias components (e.g., labeling consistency, sampling representativeness) and extend the framework to additional trust dimensions such as privacy and robustness, using subjective evidence tailored to each. Broader validation across domains is also planned.

\vspace{-1em}
\bibliographystyle{splncs04}
\bibliography{uai2025-template}

\end{document}